\documentclass[10pt,twocolumn,letterpaper]{article}

\usepackage[pagenumbers]{cvpr}      %

\ifdefined\pdfminorversion
\fi

\makeatletter
\apptocmd{\@maketitle}{\begin{minipage}{\textwidth}
    \centering
    \captionsetup{hypcap=false}
    \includegraphics[width=0.83\textwidth]{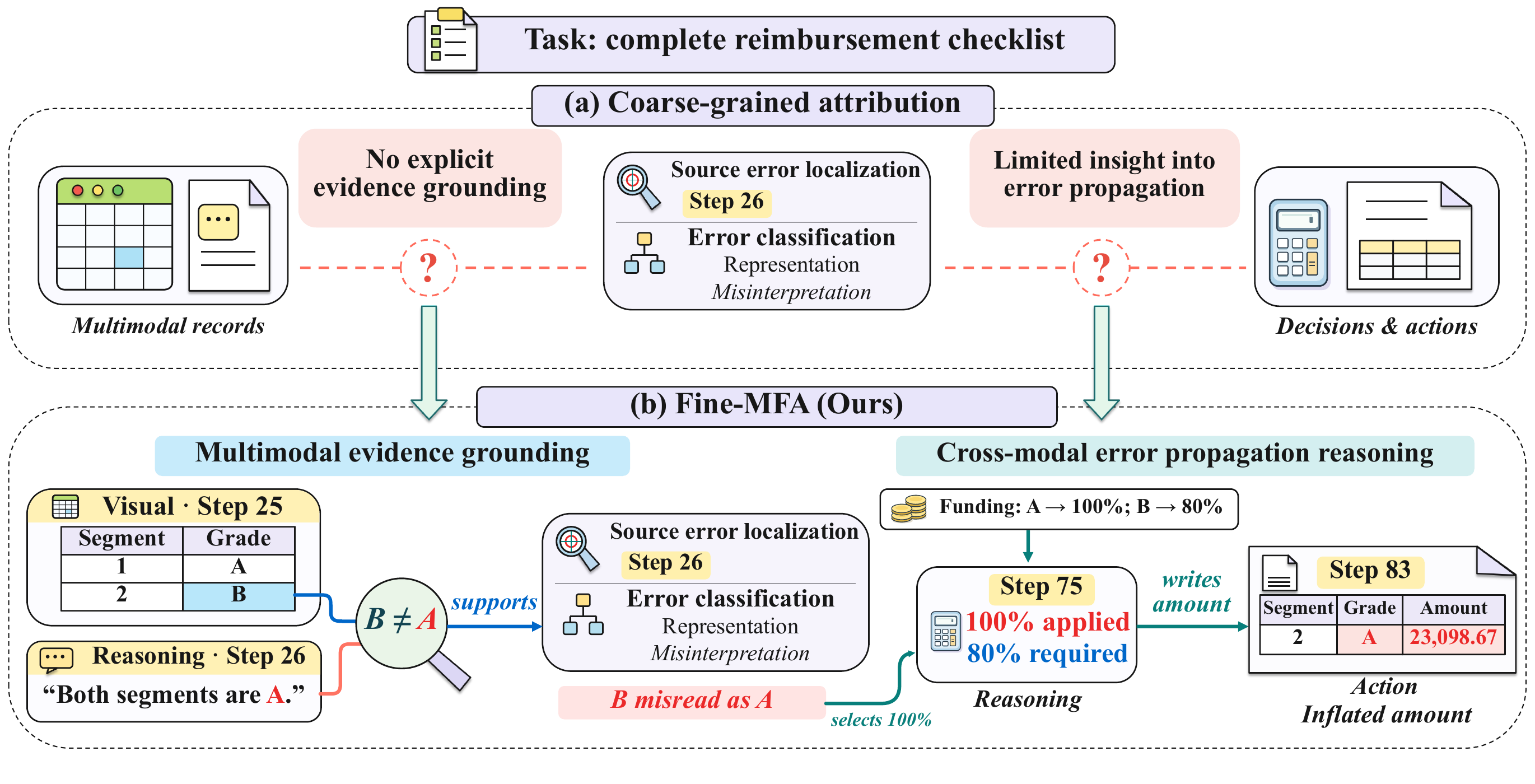}
    \captionof{figure}{From coarse-grained to fine-grained multimodal failure attribution.
While coarse-grained attribution localizes and classifies
the source error, Fine-MFA additionally grounds the diagnosis
in multimodal evidence and traces error propagation through
subsequent decisions and actions.
In this example, the agent misreads grade B
as A at step 26, leading to the application of a 100\%
allowance instead of the required 80\% at step 75 and
an inflated amount at step 83.
Only representative evidence and propagation events are shown.}
    \label{fig:overview}
\end{minipage}
\par\vspace{12pt}
}{}{
    \PackageError{fine-mfa}{Could not append the teaser to the title block}{}
}
\makeatother

\usepackage{placeins}
\usepackage{needspace}
\AtBeginDocument{\raggedbottom}

\renewcommand{\topfraction}{0.8}
\renewcommand{\dbltopfraction}{0.55}
\renewcommand{\textfraction}{0.15}
\usepackage{colortbl}
\definecolor{gainblue}{HTML}{0EBEFF}
\definecolor{lossrose}{HTML}{FF42B3}

\usepackage{inconsolata}
\usepackage{float}

\definecolor{cvprblue}{rgb}{0.21,0.49,0.74}
\usepackage[pagebackref,breaklinks,colorlinks,allcolors=cvprblue]{hyperref}

\def\confName{CVPR}
\def\confYear{2027}

\title{Fine-MFA: Towards Fine-grained Failure Attribution for Multimodal Agents}

\author{
\textbf{Weixiang Zhao}\thanks{Equal contribution.}, \textbf{Yuhang Gu}\footnotemark[1], \textbf{Xingyu Sui}, \textbf{De Hua}, \textbf{Qiming Yu},\\
\textbf{Yanyan Zhao}, \textbf{Bing Qin}, \textbf{Ting Liu}\\
Harbin Institute of Technology\\
{\fontencoding{T1}\fontfamily{zi4}\selectfont \{wxzhao, yhgu, yyzhao\}@ir.hit.edu.cn}
}

\begin{document}
\maketitle

\begin{abstract}
Understanding why multimodal agents fail is essential
for improving their reliability in complex interactive
environments.
However, existing failure attribution studies largely
focus on locating and classifying errors, overlooking
the multimodal evidence underlying failures and how
errors propagate through subsequent interactions.
This limits the verifiability of failure diagnoses
and leaves their utility for downstream recovery unclear.
To bridge these gaps, we introduce \textbf{Fine-MFA},
a unified evaluation framework for fine-grained
multimodal failure attribution.
Fine-MFA extends conventional error diagnoses with
multimodal evidence grounding and cross-modal error
propagation reasoning, supported by annotated
real-world failure trajectories from GUI and embodied
environments.
Through systematic evaluations, we reveal that current
agents struggle with fine-grained attribution,
particularly in reconstructing error propagation,
despite their ability to generate plausible,
evidence-supported explanations.
We further demonstrate the importance of visual
evidence in accurately diagnosing cross-modal failures.
Beyond attribution accuracy, we investigate whether
fine-grained diagnoses facilitate downstream recovery.
Our results reveal greater recovery benefits for
cross-modal failures, with multimodal evidence providing
particularly valuable guidance.
Further analysis shows that these benefits depend on
how agents utilize different diagnostic components,
highlighting the need for adaptive attribution utilization.
Together, our findings establish a comprehensive
perspective on multimodal failure attribution,
connecting diagnostic accuracy with practical
recovery utility.
\end{abstract}

\section{Introduction}
\label{sec:introduction}

Multimodal agents interact with visual environments by
perceiving their surroundings, reasoning about task goals,
and executing actions, enabling applications such as GUI
automation and embodied manipulation
\citep{koh2024visualwebarena,xie2024osworld,yang2025embodiedbench}.
Despite their growing capabilities, these agents remain
prone to errors during task execution, often resulting
in unsuccessful outcomes.
Analyzing and understanding such failures is essential
for identifying agent limitations and guiding targeted
improvements to their performance
\citep{cemri2025mast,zhu2025agentdebug}.

Failure attribution aims to identify the underlying causes of agent failures by tracing errors in execution trajectories \citep{zhang2025whowhen,cemri2025mast}.
This task becomes particularly challenging in multimodal environments, where agents interact through visual observations, reasoning, actions, and environmental feedback.
An observed failure may originate from an earlier misinterpretation of visual information, incorrect associations between objects and attributes, or erroneous decisions, making it difficult to identify the actual source and understand how errors propagate across steps \citep{zhang2026cuadebug,qi2026trajdebug}.

Despite recent advances, existing failure attribution studies
largely focus on coarse-grained diagnoses, such as identifying
erroneous steps or error categories
\citep{zhang2025whowhen,liu2026whowhenpro,deshpande2025trail}.
As illustrated in Figure~\ref{fig:overview}(a),
coarse-grained attribution can identify a representation-stage
misinterpretation at step 26 in a reimbursement task,
but leaves two critical questions unanswered.
First, what evidence establishes that the agent
misinterpreted the information?
Without explicit multimodal evidence grounding,
it is difficult to verify a diagnosis against visual
observations, reasoning traces, and environmental feedback
\citep{zhang2026cuadebug}.
Second, how does this initial error contribute to
the final task failure?
Without tracing its influence on subsequent decisions
and actions, the utility of such diagnoses for
downstream recovery remains unclear \citep{wang2026survey}.

To bridge these gaps, we introduce \textbf{Fine-MFA},
a unified evaluation framework for
\textbf{\underline{Fine}}-grained
\textbf{\underline{M}}ultimodal
\textbf{\underline{F}}ailure
\textbf{\underline{A}}ttribution.
Fine-MFA extends conventional error localization and
classification with multimodal evidence grounding and
cross-modal error propagation reasoning.
As shown in Figure~\ref{fig:overview}(b),
the worksheet at step 25 records grade B, whereas
the agent interprets it as A at step 26,
providing verifiable evidence for the diagnosis.
This misinterpretation subsequently leads the agent
to apply a full allowance instead of the required
reduced rate at step 75, resulting in an inflated
amount at step 83.
By connecting error sources with their supporting
evidence and downstream consequences, Fine-MFA
enables a more comprehensive characterization
of multimodal failures.
Specifically, we establish a four-dimensional
attribution taxonomy, with error classification
further distinguishing six error stages and nine
error mechanisms.
Based on this taxonomy, we collect and annotate
real-world failure trajectories from embodied and
GUI agent benchmarks
\citep{yang2025embodiedbench,xie2024osworld,yuan2026osworld2}.
Building upon this dataset, we systematically
investigate fine-grained multimodal failure attribution
from two perspectives: attribution capability and
downstream recovery utility.

We first investigate (\textbf{RQ1}) \emph{how accurately
can current agents perform fine-grained multimodal
failure attribution?} (\S\ref{sec:attribution-evaluation})
To answer this question, we evaluate five models,
each operating independently or with one of two
advanced harnesses, yielding 4,050 attribution results.
Our evaluation reveals substantial limitations in
current agents' fine-grained attribution capabilities.
In particular, agents struggle to accurately identify
error sources and mechanisms, despite their ability
to provide supporting evidence
(\S\ref{subsec:attribution-performance}).
This limitation is especially pronounced in error
propagation reasoning, where semantically supported
explanations often fail to reconstruct the underlying
propagation structure.
Although tool-augmented harnesses improve overall
attribution performance, their gains partly arise
from improved response coverage rather than
diagnostic accuracy.
Furthermore, visual ablations reveal that removing
visual evidence disproportionately impairs error
classification and propagation reasoning for
cross-modal failures
(\S\ref{subsec:visual-ablation}).
These findings highlight the challenges of fine-grained
multimodal attribution and the critical role of
multimodal evidence in reliable failure diagnosis.

These findings naturally lead to our second question:
(\textbf{RQ2}) \emph{does fine-grained attribution improve
downstream error recovery compared with coarse-grained
attribution?} (\S\ref{sec:recovery})
To answer this question, we evaluate five recovery
models using gold-standard coarse- and fine-grained
attributions, yielding 1,350 paired comparisons.
We find that fine-grained attribution provides greater
recovery benefits for cross-modal failures, yielding
an observed improvement of 2.48 percentage points,
compared with a decrease of 0.64 points on
text-sufficient failures
(\S\ref{sec:recovery-gold}).
Notably, within the same model families, variants
with weaker baseline recovery performance tend to
benefit more from fine-grained feedback.
Further component ablations reveal that multimodal
evidence provides particularly valuable recovery
guidance, while the contribution of error propagation
varies across models and feedback combinations
(\S\ref{sec:recovery-component-ablation}).
These findings suggest that fine-grained attribution
offers richer diagnostic guidance for recovery,
while realizing its full potential requires
agents to adaptively utilize the available
attribution information.

In summary, our findings reveal substantial challenges in fine-grained multimodal failure attribution, particularly in reconstructing error propagation, while highlighting the critical role of multimodal
evidence in reliable diagnosis. Furthermore, fine-grained attribution shows greater potential for recovering from cross-modal failures, although its effectiveness depends on how agents utilize the available diagnostic information. By jointly evaluating attribution capability and downstream recovery utility, \textbf{Fine-MFA} provides a unified foundation for understanding current limitations and guiding the adaptive utilization of multimodal failure attributions.

\begin{figure*}[t]
    \centering
    \captionsetup[subfigure]{
        font=small,
        justification=centering,
        singlelinecheck=false
    }

    \begin{subfigure}[t]{0.164\textwidth}
        \centering
        \includegraphics[width=\linewidth]{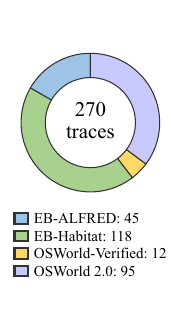}
        \caption{Data sources}
        \label{fig:profiles-sources}
    \end{subfigure}\hfill%
    \begin{subfigure}[t]{0.164\textwidth}
        \centering
        \includegraphics[width=\linewidth]{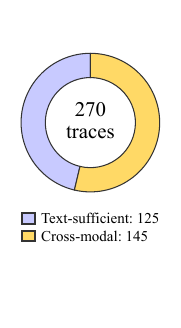}
        \caption{Evidence groups}
        \label{fig:profiles-groups}
    \end{subfigure}\hfill%
    \begin{subfigure}[t]{0.280\textwidth}
        \centering
        \includegraphics[width=\linewidth]{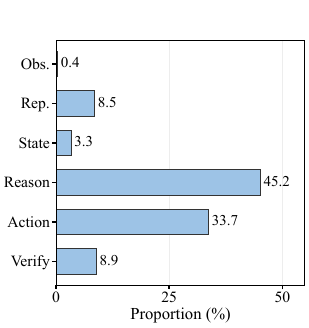}
        \caption{Source stages}
        \label{fig:profiles-stages}
    \end{subfigure}\hfill%
    \begin{subfigure}[t]{0.362\textwidth}
        \centering
        \includegraphics[width=\linewidth]{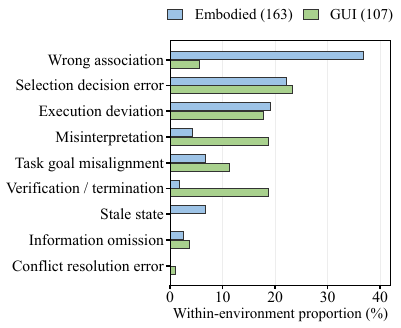}
        \caption{Error mechanisms}
        \label{fig:profiles-mechanisms}
    \end{subfigure}

    \caption{{Failure profiles in Fine-MFA.}
    (a) Data sources and (b) evidence groups partition the 270 traces.
    (c) Source-stage proportions use all traces.
    (d) Primary mechanism proportions are computed within the embodied
    ($n=163$) and GUI ($n=107$) groups.}
    \label{fig:failure-profiles}
\end{figure*}

\section{Fine-MFA: Schema and Dataset}
\label{sec:framework}

\subsection{Attribution Schema}
\label{sec:attribution-schema}

To move beyond coarse-grained failure diagnosis, Fine-MFA organizes attribution into four complementary dimensions: source error localization, error classification, multimodal evidence grounding, and cross-modal error propagation reasoning.
Together, they identify where an error originates, explain why it occurs, establish supporting evidence, and trace its consequences.

\textbf{Source error localization (SEL).}
A failed trajectory may contain multiple errors, not all
of which contribute to the final failure.
SEL therefore aims to identify the earliest consequential
error rather than simply the first observed mistake.
The source is represented as an inclusive interval of
normalized observation--decision steps, each potentially
containing multiple actions.
Errors that are subsequently corrected without affecting
the outcome are excluded.

\textbf{Error classification (EC).}
Locating an error alone does not explain its underlying
cause, especially in multimodal interactions where failures
may arise at different processing stages.
We therefore distinguish six error stages
(e.g., observation, reasoning, and action)
and nine error mechanisms
(e.g., misinterpretation, wrong association, and
execution deviation), with the complete taxonomy
detailed in the Appendix~\ref*{app:taxonomy}.
Separating stages from mechanisms allows us to distinguish
where an error emerges from how it arises, providing a more
precise characterization of diverse multimodal failures.

\textbf{Multimodal evidence grounding (MEG).}
A plausible diagnosis is not necessarily supported by
the execution record.
To ensure verifiable attribution, MEG grounds diagnostic
claims in retrievable multimodal evidence, including visual
observations, reasoning traces, and environmental feedback.
Each evidence item specifies its source, location, and
supporting fact, while multiple items may jointly
substantiate a diagnosis.

\textbf{Cross-modal error propagation reasoning (CEPR).}
Identifying an error source does not reveal how it
contributes to the eventual failure.
CEPR therefore traces how an initial error influences
subsequent reasoning, decisions, and actions.
We represent propagation as a source-anchored sequence, recording their steps, stages, and supporting evidence.
Connections are established only when supported by
the trajectory, rather than by temporal proximity alone.

\paragraph{Reimbursement checklist example.}
As illustrated in Figure~\ref{fig:overview},
the agent misreads grade B as A at step 26 (SEL),
constituting a representation-stage misinterpretation (EC).
The worksheet screenshot and reasoning record substantiate
this diagnosis (MEG).
The error then propagates to the application of the
full allowance at step 75 and an inflated entry at
step 83 (CEPR).

\subsection{Dataset and Reference Annotations}
\label{sec:fine-mfa-data}

We curate 270 real failed execution trajectories from
EmbodiedBench, OSWorld-Verified, and OSWorld~2.0
\citep{yang2025embodiedbench,xie2024osworld,yuan2026osworld2}.
To enable evidence-grounded failure analysis, we select
trajectories that preserve task instructions, visual
observations, available reasoning traces, actions,
environmental feedback, and execution outcomes.
Each trajectory represents an individual execution,
allowing multiple failed runs of the same task while
retaining their original identifiers and agent information.
As shown in Figure~\ref{fig:failure-profiles}(a),
the dataset comprises 163 embodied and 107 GUI trajectories.
These environments differ substantially in trajectory
length, with median lengths of 11 and 167 steps and
10 and 155 distinct visual frames, respectively.

To establish reliable reference attributions, two annotators independently label each trajectory across the four dimensions of Fine-MFA.
Their annotations undergo model-assisted review against the original execution records, followed by human adjudication of discrepancies.
The detailed annotation protocol and quality control procedures are provided in Appendix~\ref*{app:annotation}.
Figure~\ref{fig:failure-profiles}(c,d) illustrates
the diversity of annotated failures: source errors
predominantly occur during reasoning and action,
while error mechanisms vary across embodied and GUI
environments.

To examine when visual information is essential for failure attribution, we further divide trajectories into two evidence-dependency groups.
A trajectory is \emph{text-sufficient} if its diagnosis can be verified from textual records, actions, and structured feedback, whereas a \emph{cross-modal} trajectory requires visual evidence to establish at least one critical fact. For example, an agent's textual description of an object reflects its belief but may require visual verification to determine whether it is correct.
This distinction captures evidential necessity rather than image availability or model difficulty. As shown in Figure~\ref{fig:failure-profiles}(b), the resulting groups contain 125 text-sufficient and 145 cross-modal trajectories, established before conducting visual ablation experiments.

\section{RQ1: Evaluating Fine-Grained Multimodal Failure Attribution}
\label{sec:attribution-evaluation}

In this section, we investigate RQ1 by first introducing
the evaluation protocol and metrics
(\S\ref{subsec:attribution-protocol}).
We then assess the fine-grained attribution capabilities
of standalone models and tool-augmented harnesses
(\S\ref{subsec:attribution-performance}).
Finally, we conduct visual evidence ablations to examine
the role of multimodal information in failure attribution
(\S\ref{subsec:visual-ablation}).

\subsection{Evaluation Protocol and Metrics}
\label{subsec:attribution-protocol}

We evaluate five models (GPT-5.6 Sol
\citep{openai2026gpt56},
Kimi K2.6 \citep{moonshotai2026kimik26},
Qwen3.8-27B \citep{qwen2026qwen38},
DeepSeek-V4.1-Flash \citep{deepseekai2026deepseekv41flash},
and Claude Sonnet 5.5 \citep{anthropic2026sonnet55}),
each operating independently or with one of two advanced
harnesses, on the same 270 failed trajectories.
In the standalone (Bare) setting, each model receives the
complete trajectory, including its original images, in a
single request.
In contrast, Codex Harness and DeepSeek Harness allow
models to interactively retrieve trajectory records,
inspect visual evidence, and refine their diagnoses
through native tool loops.
This yields 4,050 attribution results, evaluated against
held-out reference annotations.

Each setting follows its native context configuration,
with a 25-minute limit for Bare and a 60-minute limit
for either harness.
The standard output budget is 32K tokens per model call. Detailed evaluation configurations are provided in
Appendix~\ref*{app:attribution-setup}.
We evaluate the resulting diagnoses along four dimensions,
using the following metrics.

\paragraph{Source error localization and classification.}
For valid predicted and reference intervals
$P=[p_s,p_e]$ and $R=[r_s,r_e]$,
we compute intersection over union by counting the steps
in each inclusive interval.
Let $d=(|p_s-r_s|+|p_e-r_e|)/2$ be the mean endpoint distance
and $w=\min(|P|,|R|)/\max(|P|,|R|)$ the width ratio.
The SEL score is
\[
    100 \times \max\{\operatorname{IoU}(P,R),\,2^{-(d+1)}w\}.
\]
A missing or invalid interval scores zero.
For stage match $S$ and primary mechanism match $M$,
each in $\{0,1\}$, EC is
$100 \times (0.25S+0.50M+0.25SM)$.
This metric prioritizes mechanism identification while
rewarding consistency between error stage and mechanism.

\paragraph{Multimodal evidence grounding and error propagation.}
MEG evaluates whether a diagnosis is supported by
reliable evidence.
It combines evidence quality $q$, measuring citation
locatability, fidelity, and diagnostic support, with
coverage $c$ of the required facts.
Evidence quality uses weights of $0.2$, $0.4$, and $0.4$
for its three components, and MEG is computed as
100 times the harmonic mean of $q$ and $c$.

CEPR evaluates both the structural accuracy and
semantic validity of reconstructed error propagation.
It combines event-node F1, directed-edge F1, and
semantic support with weights of $0.4$, $0.2$, and $0.4$.
Events are matched by step and stage, while edges
must connect matched events in the correct direction.
Semantic judgments assess whether the trajectory
supports the predicted propagation links.
These judgments are assisted by GPT-5.6 Sol,
with model-assisted review for adjudication
and required-fact coverage.

\paragraph{Overall score and response coverage.}
To capture overall attribution performance while
preserving the distinctions among dimensions,
we report SEL, EC, MEG, and CEPR separately,
alongside the Hierarchical Attribution Score (HAS).
Let $N$ denote the number of evaluation trajectories,
and $L_i$, $C_i$, $E_i$, and $P_i$ denote the four scores
for trajectory $i$, respectively.
HAS aggregates these scores with a per-response
format factor $f_i$:
\[
    \overline{\mathrm{HAS}} =
    \frac{1}{N}\sum_{i=1}^{N}
    f_i(0.3L_i+0.3C_i+0.2E_i+0.2P_i).
\]
The factor is 1.00 for compliant output,
0.95 for minor formatting defects,
0.85 for recoverable structural defects,
or 0.70 for free text with extractable attribution.
Empty and process-only responses receive zero.
All scores are averaged over the full evaluation set,
jointly reflecting attribution quality and response coverage.
Detailed metric definitions and scoring procedures
are provided in Appendix~\ref*{app:scoring}.

\begin{table}[!t]
\centering\small
\caption{\textbf{Attribution results on 270 trajectories per setting.} Codex and DeepSeek denote harnesses. Scores include empty outputs; bold marks the best HAS within each model.}
\label{tab:attribution-main}
\setlength{\tabcolsep}{3pt}
\renewcommand{\arraystretch}{1.05}
\begin{tabular*}{\columnwidth}{@{\extracolsep{\fill}}lrrrrr@{}}
\toprule
& \multicolumn{4}{c}{\textbf{Component scores}} & \textbf{Overall}\\
\cmidrule(lr){2-5}\cmidrule(l){6-6}
\textbf{Setting} & SEL $\uparrow$ & EC $\uparrow$ & MEG $\uparrow$ & CEPR $\uparrow$ & \textbf{HAS} $\uparrow$\\
\midrule
\rowcolor{gray!7}
\multicolumn{6}{c}{\textit{GPT-5.6 Sol}}\\
Bare & 36.46 & 30.56 & 60.97 & 25.22 & 37.20 \\
Codex & 39.26 & 35.56 & 75.71 & 40.57 & \textbf{45.51} \\
DeepSeek & 39.83 & 35.00 & 73.24 & 33.14 & 43.41 \\
\midrule[0.3pt]
\rowcolor{gray!7}
\multicolumn{6}{c}{\textit{Kimi K2.6}}\\
Bare & 15.87 & 14.35 & 8.21 & 4.51 & 9.47 \\
Codex & 33.14 & 24.44 & 40.83 & 21.19 & \textbf{29.23} \\
DeepSeek & 31.04 & 26.39 & 34.77 & 18.55 & 27.07 \\
\midrule[0.3pt]
\rowcolor{gray!7}
\multicolumn{6}{c}{\textit{Qwen3.8-27B}}\\
Bare & 27.08 & 19.07 & 38.84 & 13.70 & 23.63 \\
Codex & 35.92 & 31.67 & 74.10 & 26.65 & \textbf{39.36} \\
DeepSeek & 31.65 & 25.56 & 59.61 & 21.98 & 32.58 \\
\midrule[0.3pt]
\rowcolor{gray!7}
\multicolumn{6}{c}{\textit{DeepSeek-V4.1-Flash}}\\
Bare & 20.66 & 14.17 & 26.61 & 13.19 & 17.24 \\
Codex & 41.10 & 31.30 & 67.12 & 28.67 & \textbf{38.81} \\
DeepSeek & 33.01 & 27.22 & 53.29 & 20.70 & 31.10 \\
\midrule[0.3pt]
\rowcolor{gray!7}
\multicolumn{6}{c}{\textit{Claude Sonnet 5.5}}\\
Bare & 27.83 & 19.44 & 31.71 & 10.42 & 21.10 \\
Codex & 36.33 & 25.56 & 56.30 & 21.08 & 33.21 \\
DeepSeek & 35.96 & 26.39 & 58.96 & 23.59 & \textbf{34.64} \\
\bottomrule
\end{tabular*}
\end{table}

\begin{figure*}[t]
    \centering
    \captionsetup{hypcap=false}
    \includegraphics[width=\textwidth]{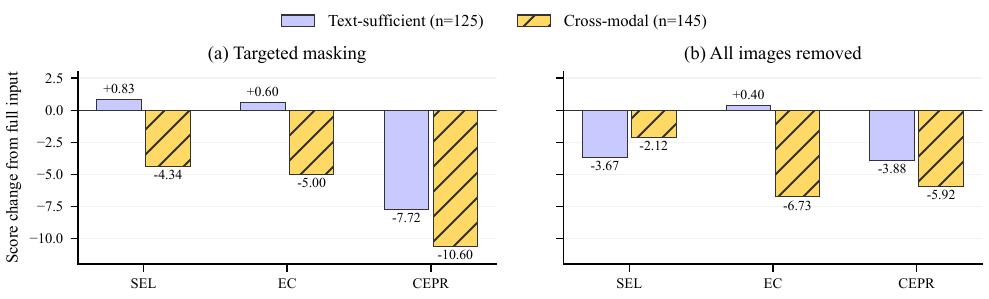}
    \caption{{Visual evidence ablation with GPT-5.6 Sol and Codex Harness.} Changes in SEL, EC, and CEPR relative to full input under targeted masking (left) and removal of all images (right). Colors distinguish text-sufficient ($n=125$) and cross-modal ($n=145$) traces. Negative values indicate lower scores on the 0--100 scale.}
    \label{fig:visual-evidence}
\end{figure*}

\subsection{Attribution Performance}
\label{subsec:attribution-performance}

Table~\ref{tab:attribution-main} presents the attribution
performance of five models across standalone and
tool-augmented settings.
We first examine their fine-grained attribution
capabilities and limitations, followed by the impact
of tool-augmented harnesses.

\textbf{Current agents struggle with fine-grained
failure attribution.}
Even the best-performing configuration, GPT-5.6 Sol
with Codex Harness, achieves a HAS of only 45.51.
In particular, it scores 39.26 on source error
localization and 35.56 on error classification,
despite achieving 75.71 on evidence grounding.
These results reveal a substantial gap between
providing supporting evidence and accurately
identifying the source and mechanism of failures.

\textbf{Error propagation reasoning remains
particularly challenging.}
For GPT-5.6 Sol with Codex, propagation explanations
achieve 84.92 in semantic support, whereas event-node
and directed-edge F1 reach only 15.04 and 2.91,
respectively.
This discrepancy indicates that agents can generate
evidence-supported explanations but struggle to
reconstruct the precise structure of error propagation.

\textbf{Tool-augmented harnesses improve overall
performance, but their gains vary in origin.}
Both harnesses outperform standalone inference
across all five models, with Codex Harness achieving
the highest HAS for four models.
However, these improvements reflect both response
coverage and attribution quality.
For GPT-5.6 Sol, the 8.31-point HAS improvement with Codex
Harness primarily stems from increased response
coverage, with negligible change on commonly
answered trajectories.
In contrast, Kimi improves by 21.23 points on
commonly answered trajectories, demonstrating
substantial gains in attribution quality.

\subsection{Role of Visual Evidence}
\label{subsec:visual-ablation}

To examine the role of visual evidence in failure
attribution, we conduct two ablations using Sol with
Codex Harness: targeted masking of reference-relevant
images and complete removal of visual observations.
Both settings preserve textual records, actions, and
environmental feedback.
Figure~\ref{fig:visual-evidence} compares the resulting
changes in SEL, EC, and CEPR across text-sufficient
and cross-modal trajectories.
We draw the following key observations.

\begin{table*}[t]
\centering\small
\caption{\textbf{Recovery with coarse and fine-grained gold feedback.} Rates are percentages; $\Delta$ is Fine minus Coarse in percentage points. \colorbox{gainblue!45}{\strut blue} cells indicate improvement, while \colorbox{lossrose!35}{\strut pink} cells indicate decline; darker shades denote larger changes. Bold marks the highest model-level $\Delta$ in each subset.}
\label{tab:recovery-main}
\setlength{\tabcolsep}{4pt}
\renewcommand{\arraystretch}{1.1}
\begin{tabular*}{\textwidth}{@{\extracolsep{\fill}}lrrrrrrrrr@{}}
\toprule
& \multicolumn{3}{c}{\textbf{All traces} ($n=270$)} & \multicolumn{3}{c}{\textbf{Text-sufficient} ($n=125$)} & \multicolumn{3}{c}{\textbf{Cross-modal} ($n=145$)}\\
\cmidrule(lr){2-4}\cmidrule(lr){5-7}\cmidrule(l){8-10}
\textbf{Model} & Coarse & Fine & $\Delta$ & Coarse & Fine & $\Delta$ & Coarse & Fine & $\Delta$\\
\midrule
GPT-5.6 Sol & 37.78 & 38.15 & \cellcolor{gainblue!15.7} +0.37 & 33.60 & 33.60 & 0.00 & 41.38 & 42.07 & \cellcolor{gainblue!18.9} +0.69 \\
GPT-5.6 Luna & 32.96 & 36.30 & \cellcolor{gainblue!45.3} \textbf{+3.33} & 30.40 & 31.20 & \cellcolor{gainblue!20.0} \textbf{+0.80} & 35.17 & 40.69 & \cellcolor{gainblue!67.2} \textbf{+5.52} \\
Qwen3.5-9B & 26.67 & 27.41 & \cellcolor{gainblue!19.4} +0.74 & 25.60 & 24.00 & \cellcolor{lossrose!28.0} -1.60 & 27.59 & 30.34 & \cellcolor{gainblue!39.6} +2.76 \\
Qwen3.5-27B & 24.44 & 23.70 & \cellcolor{lossrose!19.4} -0.74 & 20.00 & 18.40 & \cellcolor{lossrose!28.0} -1.60 & 28.28 & 28.28 & 0.00 \\
Claude Sonnet 5.5 & 32.59 & 34.07 & \cellcolor{gainblue!26.8} +1.48 & 33.60 & 32.80 & \cellcolor{lossrose!20.0} -0.80 & 31.72 & 35.17 & \cellcolor{gainblue!46.5} +3.45 \\
\midrule
\rowcolor{gray!8}
\textbf{All models} & 30.89 & 31.93 & \cellcolor{gainblue!22.4} +1.04 & 28.64 & 28.00 & \cellcolor{lossrose!18.4} -0.64 & 32.83 & 35.31 & \cellcolor{gainblue!36.8} +2.48 \\
\bottomrule
\end{tabular*}
\end{table*}

\textbf{Visual evidence is particularly important
for identifying cross-modal failure mechanisms.}
Removing all images reduces EC by 6.73 points
on cross-modal trajectories, compared with a
0.40-point increase on text-sufficient trajectories.
Targeted masking exhibits a similar pattern,
with changes of $-5.00$ and $+0.60$ points,
respectively.
These results demonstrate that visual information
plays a critical role in distinguishing failure
mechanisms that cannot be verified from textual
records alone.

\textbf{Error propagation reasoning is sensitive
to the removal of relevant visual evidence.}
As shown in Figure~\ref{fig:visual-evidence},
targeted masking reduces the CEPR score
from 40.57 to 31.30, compared with 35.59
under complete image removal.
This suggests that removing diagnostically relevant
visual observations can substantially affect
propagation, even when other visual
information remains available.

\textbf{Source error localization depends on
the availability of relevant visual evidence.} Targeted masking reduces SEL by 4.34 points on
cross-modal trajectories, while increasing it by 0.83 points on text-sufficient trajectories.
In contrast, removing all images leads to decreases
of 2.12 and 3.67 points, respectively.
These results suggest that source localization depends
not only on the availability of visual information,
but also on its relevance to the underlying failure.

\textbf{Summary.}
Overall, our findings reveal substantial limitations
in current agents' fine-grained multimodal failure
attribution capabilities, particularly in reconstructing
error propagation.
Although tool-augmented harnesses improve overall
attribution performance, these gains partly reflect
increased response coverage rather than consistently
better diagnostic accuracy.
Moreover, agents often generate semantically supported
explanations without accurately reconstructing the
underlying propagation structure, revealing a persistent
gap between plausible reasoning and precise attribution.
Visual ablations further demonstrate that multimodal
evidence plays a critical role in accurate failure
diagnosis, especially when identifying error mechanisms
and tracing their propagation across modalities.
The greater sensitivity of cross-modal failures
to visual evidence removal further highlights the
limitations of relying on textual execution records
alone.
These findings underscore the importance of grounding
failure attribution in multimodal observations
and evaluating diagnostic capabilities beyond
surface-level explanations.

\section{RQ2: Evaluating the Recovery Utility of Fine-Grained Attribution}
\label{sec:recovery}

In this section, we investigate whether fine-grained
attribution improves downstream error recovery.
We first introduce the recovery protocol
(\S\ref{sec:recovery-protocol}) and compare coarse-
and fine-grained gold attributions
(\S\ref{sec:recovery-gold}).
We then examine the individual and combined
contributions of multimodal evidence and error
propagation through component ablations
(\S\ref{sec:recovery-component-ablation}).
Finally, we analyze representative recovery failures
to understand the challenges of translating
fine-grained diagnoses into successful actions
(\S\ref{sec:recovery-error-analysis}).

\subsection{Recovery Protocol}
\label{sec:recovery-protocol}

To evaluate the practical utility of failure attribution,
we conduct retrospective recovery in which
agents resume failed tasks from a shared takeover point
immediately before the gold source decision.
For each model--trajectory pair, different feedback
conditions share the same recovery model, decoding settings,
and initial environment state.
The agent receives the task, current observation,
permitted interaction history, action interface,
and condition-specific diagnostic feedback.
Recovery success is assessed using the original task
evaluator and its predefined success criterion.

Each recovery attempt has a two-hour time limit,
with budgets of 100 actions and 120 decisions for GUI
tasks, and 50 actions and 60 decisions for embodied tasks.
The final state is evaluated when the agent terminates
or exhausts its budget.
As this is a retrospective evaluation, gold diagnostic
feedback may include evidence from the original
failed execution after the takeover point.
Detailed recovery settings are provided in
Appendix~\ref*{app:recovery-setup}.

\subsection{Recovery Performance with Gold Attributions}
\label{sec:recovery-gold}

To isolate the value of attribution granularity,
we compare two gold-feedback conditions.
\emph{Coarse} provides the source error and its
classification, while \emph{Fine} additionally includes
multimodal evidence and error propagation information.
We evaluate five recovery models on 270 failed trajectories,
yielding 1,350 paired comparisons.
Table~\ref{tab:recovery-main} presents the results,
from which we draw three key findings.

\textbf{Fine-grained attribution offers greater
recovery benefits for cross-modal failures.}
Across all five models, Fine achieves an observed
recovery improvement of 2.48 percentage points on
cross-modal trajectories, compared with a decrease
of 0.64 points on text-sufficient trajectories.
The pooled recovery rate increases from 32.83\% to
35.31\% on cross-modal failures.
This contrast suggests that additional diagnostic
evidence and propagation information are particularly
valuable when failures require reasoning across
visual and textual observations.

\textbf{Fine-grained attribution offers greater
recovery benefits to less capable agents.}
As shown in Table~\ref{tab:recovery-main},
within both the GPT-5.6 and Qwen3.5 model families,
the variants with lower baseline recovery performance
benefit more from fine-grained attribution on
cross-modal trajectories.
This suggests that richer diagnostic feedback may
help compensate for limited recovery capabilities,
particularly when failures require cross-modal reasoning.

\textbf{Recovery benefits vary across models
and do not necessarily align with baseline performance.}
Fine improves cross-modal recovery for four of the
five models.
Notably, GPT-5.6 Sol achieves the highest cross-modal
recovery rate under both conditions, but its improvement
is only 0.69 percentage points, compared with 5.52
points for GPT-5.6 Luna.
These results indicate that the ability to recover
from failures and the ability to exploit richer
attribution feedback are distinct capabilities.

\begin{figure}[!htbp]
    \centering
    \includegraphics[width=\columnwidth]{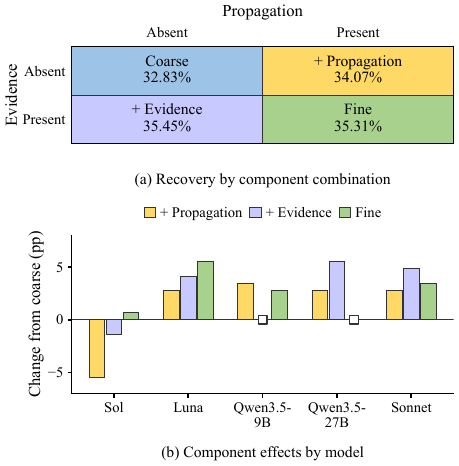}
    \caption{Evidence and propagation contributions to recovery.
    Top: pooled recovery success rates on 725 cross-modal
    model--trace pairs.
    Bottom: changes from coarse feedback on 145 cross-modal
    trajectories per model, in percentage points.
    Open squares indicate zero change.}
    \label{fig:component-effects}
\end{figure}

\subsection{Contributions of Evidence and Propagation}
\label{sec:recovery-component-ablation}

To further understand where the recovery benefits originate, we conduct component ablations on the
145 cross-modal trajectories.
Beyond Coarse and Fine, we introduce two intermediate
conditions: \emph{Coarse + Evidence}, which adds the
gold multimodal evidence and its associated materials,
and \emph{Coarse + Propagation}, which adds the gold
error propagation.
Figure~\ref{fig:component-effects} compares the four
conditions across five recovery models. We obtain the following findings.

\textbf{Multimodal evidence provides the
largest pooled recovery benefit.}
Coarse + Evidence achieves the highest pooled recovery
rate, slightly exceeding Fine.
Across the five models, adding evidence improves recovery
by 2.62 percentage points without propagation and
1.24 points with propagation.
In contrast, adding propagation yields a 1.24-point
gain without evidence but a 0.14-point decrease when
evidence is already present.
These results highlight multimodal evidence as a
particularly useful component of diagnostic feedback
for cross-modal recovery.

\textbf{The utility of error propagation
depends on its interaction with evidence and the
recovery model.}
For GPT-5.6 Sol, either component alone reduces
recovery relative to Coarse, whereas combining them
improves recovery from 41.38\% to 42.07\%.
Conversely, Qwen3.5-27B improves from 28.28\% to
33.79\% with evidence alone, despite showing no
gain from complete Fine feedback.
These contrasting results demonstrate that the individual contributions of evidence and propagation
can be obscured by coarse--fine comparisons alone, suggesting that effective recovery depends on how diagnostic information is combined and utilized.

\textbf{Effective recovery requires adaptive
utilization of fine-grained attribution.}
Fine-grained attribution generally offers greater
recovery potential than coarse-grained diagnosis,
yet providing all diagnostic components does not
always yield the best performance.
Different models benefit from different combinations
of multimodal evidence and error propagation,
suggesting that recovery depends not only on the
richness of attribution but also on how effectively
agents utilize it.
These findings highlight the importance of adaptive
attribution utilization, where agents selectively
leverage diagnostic information according to their
capabilities and the characteristics of failures.

\textbf{Summary.}
Our findings highlight the value of fine-grained attribution for cross-modal error recovery, with multimodal evidence providing particularly effective and grounded diagnostic guidance. However, the varying contributions of evidence and error propagation indicate that providing richer information alone does not guarantee better recovery. These findings suggest that fine-grained attribution provides a richer foundation for recovery, while adaptive utilization of diagnostic information is essential to fully realizing its potential.

\subsection{Error Analysis}
\label{sec:recovery-error-analysis}

To further understand the limitations of
diagnosis-guided recovery, we examine two representative
failures under fine-grained gold attribution feedback.

\textbf{Failure to act on the diagnosis.}
In a conditional transfer task involving an initially
closed refrigerator, the agent was required to move a plum.
The gold attribution identified the erroneous selection
of a plate as the source error.
However, Luna still opened the refrigerator, moved
the plate to the left counter, and terminated without
transferring the plum.
Although the environment confirmed successful action
execution, the task remained incomplete.
This case illustrates a failure to translate accurate
diagnostic feedback into appropriate actions.

\textbf{Failure after correcting the diagnosed error.}
In a can-placement task, the gold attribution identified
repeated pickup attempts at an incorrect location.
Qwen3.5-27B successfully addressed this issue by
navigating to another table and retrieving the can.
However, it subsequently placed the can in the left
drawer rather than on the required counter, causing
the task to fail despite resolving the initial obstacle.
This case highlights how new decision errors can
undermine recovery even after the diagnosed error
has been corrected.

\section{Related Work}
\label{sec:related}

\subsection{Evaluating Multimodal Agents}
Existing benchmarks evaluate agent capabilities across
diverse tasks and environments.
AgentBench, GAIA, and AgentBoard assess general agent
capabilities, including reasoning, tool use, and task progress
\citep{liu2024agentbench,mialon2024gaia,ma2024agentboard}.
WebArena, Mind2Web, and VisualWebArena evaluate web
interaction and visually grounded tasks
\citep{zhou2024webarena,deng2023mind2web,koh2024visualwebarena}.
OSWorld and AndroidWorld focus on desktop and mobile
environments \citep{xie2024osworld,rawles2025androidworld},
while EmbodiedBench and BEHAVIOR-1K assess embodied
agent capabilities \citep{yang2025embodiedbench,li2023behavior}.
$\tau$-bench further evaluates tool use in interactive
scenarios \citep{yao2024tau}.
While these benchmarks primarily assess task execution
performance, Fine-MFA shifts the focus to failure attribution,
evaluating multimodal evidence grounding and error
propagation beyond error localization and classification.

\subsection{Failure Attribution and Trace Diagnosis}
Recent studies investigate the underlying causes of agent failures \citep{chen2026seeing,zhang2026longrca,rafi2026falat,liao2026error,guo2026blade,yeh2026tracing,wang2026dependable,wang2026triage}.
MAST introduces a failure taxonomy for multi-agent systems,
while TRAIL provides annotated traces for systematic diagnosis
\citep{cemri2025mast,deshpande2025trail}.
Who\&When evaluates responsible-agent and error-step identification,
with Who\&When Pro extending attribution to diverse tasks and modalities
\citep{zhang2025whowhen,liu2026whowhenpro}.
AgenTracer and AEGIS explore scalable attribution through
counterfactual replay, error injection, and structured annotations
\citep{zhang2026agentracer,kong2026aegis}.
TRAJDEBUG and DeFA further examine error lifecycles,
event dependencies, and downstream propagation
\citep{qi2026trajdebug,deng2026defa}.
While these studies advance failure diagnosis,
Fine-MFA focuses on systematically evaluating fine-grained
attribution, extending localization and classification
with multimodal evidence grounding and cross-modal error propagation
reconstruction.

\subsection{Diagnosis-Guided Recovery}
Recent studies explore how failure diagnosis can guide recovery
\citep{zhao2024expel,zhu2026agentdebugx,wang2026darc,zhao2026debugging,zhao2026self,ni2026trace2skill}.
AgentDebug provides root-cause analysis and corrective
feedback, while CUADebug extends this approach to
visually grounded computer-use tasks
\citep{zhu2025agentdebug,zhang2026cuadebug}.
TRAJDEBUG and DeFA further investigate error propagation
and its implications for agent improvement
\citep{qi2026trajdebug,deng2026defa}.
However, the individual contributions of diagnostic
components to recovery remain underexplored.
Fine-MFA addresses this through controlled
experiments that isolate the effects of multimodal
evidence and error propagation on downstream recovery.

\section{Conclusion}
\label{sec:conclusion}

In this paper, we introduced \textbf{Fine-MFA}, a unified
evaluation framework for fine-grained multimodal failure
attribution and its downstream recovery utility.
By incorporating multimodal evidence grounding and
cross-modal error propagation reasoning, Fine-MFA
enables a systematic evaluation of failure diagnoses
beyond conventional localization and classification.
Our findings reveal that current agents struggle with
fine-grained attribution, particularly in reconstructing
error propagation, while highlighting the critical
role of multimodal evidence in reliable diagnosis.
Furthermore, controlled recovery experiments show
that fine-grained attribution offers greater benefits
for cross-modal failures, with multimodal evidence
providing particularly valuable guidance.
However, these benefits depend on how agents utilize
different diagnostic components, highlighting the
importance of adaptive attribution utilization.
Overall, Fine-MFA bridges attribution accuracy and
recovery utility, providing insights into developing
agents that can not only diagnose multimodal failures
but also effectively leverage diagnostic information
for recovery.

\section*{LLM Usage Statement}

The authors used OpenAI's ChatGPT and Codex solely for assistive purposes in preparing this manuscript. Specifically, these tools were used to assist in editing and refining author-written text for improved language and clarity, as well as in refining code and implementation. They were not used to generate related work or citations, design or analyze experiments, produce figures or visual materials, create or annotate synthetic data, or develop research ideas, hypotheses, claims, or conclusions. No full manuscript or substantial portion thereof was submitted to an LLM to obtain reviewer-style feedback. All research ideas, methodologies, experimental designs, results, and scientific conclusions were developed by the authors. The authors have reviewed and take full responsibility for the accuracy, integrity, and entirety of the submitted content.

{
    \small
    \bibliographystyle{ieeenat_fullname}
    \bibliography{main}
}

\clearpage
\makeatletter
\setlength{\@fptop}{0pt}
\makeatother
\appendix
\setcounter{table}{0}
\setcounter{figure}{0}
\renewcommand{\theHtable}{supp.\arabic{table}}
\renewcommand{\theHfigure}{supp.\arabic{figure}}
\renewcommand{\thetable}{S\arabic{table}}
\renewcommand{\thefigure}{S\arabic{figure}}
\raggedbottom
\setcounter{dbltopnumber}{3}
\renewcommand{\dbltopfraction}{0.95}
\setcounter{topnumber}{3}
\renewcommand{\topfraction}{0.9}
\renewcommand{\textfraction}{0.1}
\section{Dataset and Annotation}
\label{app:data}

\subsection{Complete Attribution Schema}
\label{app:schema}
\begin{figure*}[t]
\centering
\includegraphics[width=\textwidth]{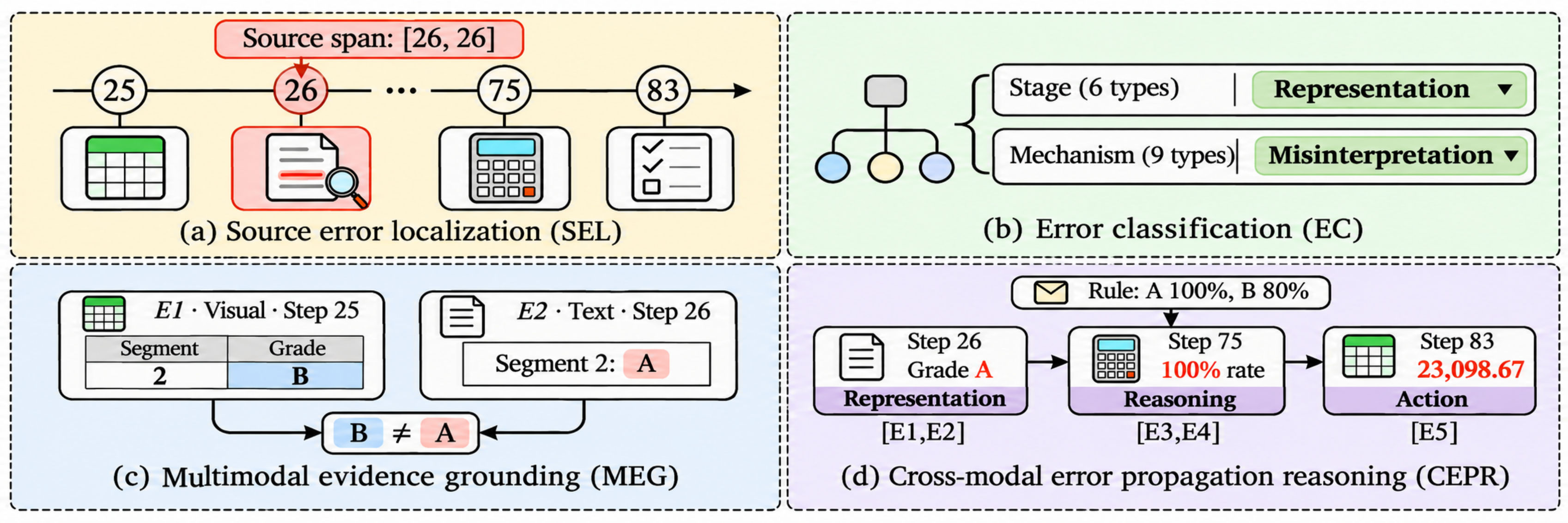}
\caption{\textbf{The four attribution components.} The source span identifies the grade misread, the classification describes its stage and mechanism, the evidence contrasts the observed and recorded grades, and the propagation connects the misread to later reasoning and action. Evidence identifiers in this schematic are local to the illustration.}
\label{fig:app-schema}
\end{figure*}

Fine-MFA uses an individual failed execution as its annotation unit. The 270 trajectories comprise 45 EB-ALFRED, 118 EB-Habitat, 12 OSWorld-Verified, and 95 OSWorld 2.0 executions. Each reference connects a source event to its classification, supporting records, and subsequent propagation. Figure~\ref{fig:app-schema} illustrates these relationships using the reimbursement example.

\paragraph{Source error event.}
The source\_error\_event object contains an inclusive step\_span, an error\_stage, and an event\_summary. The span records its start and end steps, with equal endpoints for a single-step source. An optional target\_ref identifies the specific record associated with the event. The source is the earliest error supported as contributing to the recorded failure.

\paragraph{Error mechanism.}
The error\_mechanism object assigns a primary mechanism to the source. The stage describes where the deviation occurs in processing, while the mechanism describes how it occurs. Some references also retain a secondary mechanism; EC evaluates the primary mechanism and source stage.

\paragraph{Attribution evidence.}
The attribution\_evidence list contains records with an evidence\_id, step\_id, asset\_ref, modality, evidence\_layer, role, and evidence\_description. A record\_ref can locate a structured entry, and a spatial\_anchor can identify a region within an image. The description states the fact established by the record; the role specifies how it relates to the diagnosis, for example as reference information, erroneous output, supporting evidence, or contradictory evidence. Several records may jointly establish a diagnostic fact.

\paragraph{Error propagation.}
The error\_propagation object records error\_events, first\_manifestation, terminal\_failure, and completeness. Each propagation event has a step, stage, summary, and evidence references. The source anchors the ordered event sequence; links express supported influence between successive listed events rather than adjacency in the raw trajectory. First manifestation and terminal failure provide additional context for when the source becomes observable and how the execution ends.

\paragraph{Structured example.}
The following excerpt shows selected fields from an embodied reference whose source is a reasoning-stage selection error at step 2. The task requires a wet sponge in a bathtub; the selected plan operates a separate sink faucet while leaving the sponge in the bathtub. The excerpt preserves the stored keys and values shown here but omits descriptions and other evidence items for space. Evidence references connect propagation events to items in the evidence list; they are local identifiers, not file paths.
\par\medskip
{\footnotesize\rmfamily
\begin{tabular}{@{}l@{}}
source\_error\_event:\\
\quad step\_span: \{start\_step: 2, end\_step: 2\}\\
\quad error\_stage: reasoning\\
error\_mechanism:\\
\quad primary: selection\_decision\_error\\
attribution\_evidence: [\ldots]\\
\quad evidence\_id: evidence\_005\\
\quad step\_id: 2\\
\quad asset\_ref: trajectory.json\\
\quad modality: text\\
\quad evidence\_layer: environment\_feedback\\
\quad role: supporting\\
error\_propagation:\\
\quad error\_events: [\ldots]\\
\qquad step\_id: 3\\
\qquad error\_stage: reasoning\\
\qquad evidence\_refs: [evidence\_005]\\
\quad first\_manifestation: \{step\_id: 2, \ldots\}\\
\quad terminal\_failure: \{step\_id: 11, \ldots\}\\
\quad completeness: complete
\end{tabular}
}
\par\medskip
The supporting feedback records a failed pickup while the sponge remains in the bathtub. The step-3 propagation event describes the subsequent focus on reachability and faucet state instead of correcting the missing wetting operation. The step-3 event cites evidence\_005 from step 2: the failed pickup is the feedback available when the agent next reasons about the faucet.

\subsection{Error Taxonomy}
\label{app:taxonomy}
For each trajectory, annotators locate the earliest error that contributed to the recorded failure and assign it two labels.
The source stage identifies where the error arose; the primary mechanism describes how it arose.
In the reimbursement example, the worksheet shows grade B for the second segment, but the agent describes that segment as grade A.
The source stage is \emph{representation} because the error appears in the agent's description of the worksheet.
The primary mechanism is \emph{misinterpretation} because the agent identifies the correct segment but misreads its grade.
Tables~\ref{tab:app-stages} and~\ref{tab:app-mechanisms} define the six stages and nine primary mechanisms, respectively.

\begin{table*}[!t]
\centering
\small
\caption{Six source stages: where the source error occurs.}
\label{tab:app-stages}
\renewcommand{\arraystretch}{1.15}
\begin{tabular}{@{}p{0.27\textwidth}p{0.69\textwidth}@{}}
\toprule
\textbf{Stage} & \textbf{Definition} \\
\midrule
\textbf{Observation} & Acquiring or reading the available observation, including visual or textual facts. \\ \addlinespace[2pt]
\textbf{Representation} & Converting observations into a description, association, or other derived representation. \\ \addlinespace[2pt]
\textbf{Internal state} & Retaining, retrieving, or updating information used across steps. \\ \addlinespace[2pt]
\textbf{Reasoning} & Drawing conclusions, planning, or choosing actions from the available information and task goal. \\ \addlinespace[2pt]
\textbf{Action} & Executing an interaction with the environment. \\ \addlinespace[2pt]
\textbf{Verification} & Checking progress, validating results, or deciding that execution should terminate. \\ \addlinespace[2pt]
\bottomrule
\end{tabular}
\end{table*}

Table~\ref{tab:app-boundaries} illustrates three distinctions used to assign a source label.
\begin{table}[htbp]
\centering\small
\caption{Illustrative decisions at three taxonomy boundaries.}
\label{tab:app-boundaries}
\begin{tabular}{@{}p{.30\columnwidth}p{.61\columnwidth}@{}}
\toprule
Distinction & Diagnostic question \\
\midrule
Representation / internal state & Was the description wrong when formed, or did a correct description become stale after an update? \\
Misinterpretation / wrong association & Did the agent misread the value, or read it correctly and assign it to another entity? \\
Selection / execution deviation & Did the agent choose the wrong target, or choose the correct target and act on a different one? \\
\bottomrule
\end{tabular}
\end{table}

\begin{table*}[!t]
\centering
\small
\caption{Nine primary error mechanisms: how the source error occurs.}
\label{tab:app-mechanisms}
\renewcommand{\arraystretch}{1.15}
\begin{tabular}{@{}p{0.27\textwidth}p{0.69\textwidth}@{}}
\toprule
\textbf{Mechanism} & \textbf{Definition} \\
\midrule
\textbf{Information omission} & Relevant information is left out of the diagnosis-driving representation or decision. \\ \addlinespace[2pt]
\textbf{Misinterpretation} & The meaning or value of a recorded fact is interpreted incorrectly. \\ \addlinespace[2pt]
\textbf{Wrong association} & Information is assigned to the wrong object, entity, attribute, or relation. \\ \addlinespace[2pt]
\textbf{Stale state} & A decision uses information that no longer reflects the current state. \\ \addlinespace[2pt]
\textbf{Conflict resolution error} & Inconsistent information is reconciled incorrectly. \\ \addlinespace[2pt]
\textbf{Task goal misalignment} & The pursued objective differs from the task requirement. \\ \addlinespace[2pt]
\textbf{Selection decision error} & An unsuitable option or action is selected for the intended goal. \\ \addlinespace[2pt]
\textbf{Execution deviation} & The executed operation deviates from the intended interaction. \\ \addlinespace[2pt]
\textbf{Verification and termination error} & Progress or completion is incorrectly verified, or execution is terminated on that basis. \\ \addlinespace[2pt]
\bottomrule
\end{tabular}
\end{table*}

The primary label describes the deviation most directly supported at the source. In particular, a wrong value is distinguished from a correct value attached to the wrong entity; pursuing the wrong goal is distinguished from selecting a poor action for the correct goal.

\subsection{Annotation Protocol}
\label{app:annotation}
Two annotators independently examine each trajectory and annotate all four components. The available materials include the task instructions, observations, recorded reasoning where available, actions, environmental feedback, and execution outcome. Each annotator identifies the source interval, assigns the source stage and mechanism, cites supporting records, and describes propagation through later events.

A model reviews the two independent annotations against the recorded execution. Both annotators then discuss their source intervals, labels, evidence, and propagation accounts in light of that review and agree on one reference for the trajectory.

\subsection{Evidence-Group Assignment}
\label{app:evidence-groups}
The groups concern the information necessary to verify the reference diagnosis. A trajectory is text-sufficient when recorded text, actions, and structured feedback establish the facts needed for that verification. It is cross-modal when at least one necessary fact requires visual evidence interpreted in the task or execution context. A trajectory remains text-sufficient when its images corroborate facts already established in text.

The two annotators independently assign these labels, followed by model-assisted review and human discussion to establish the final groups before visual ablation. The split contains 125 text-sufficient and 145 cross-modal trajectories. Figure~\ref{fig:app-schema} illustrates the distinction. The grade entered by the agent establishes what the agent recorded. To diagnose a misread, the annotator also needs the grade shown on the reimbursement form. If the form image is the only record of that grade, the diagnosis requires visual evidence even when every subsequent action is logged in text. If a structured field independently records the grade on the form, the same discrepancy can be verified from the two recorded values.

Table~\ref{tab:app-source-groups} counts each trajectory once. All 12 OSWorld-Verified cases in this collection require visual evidence, whereas the other three sources contribute cases to both groups.
\begin{table}[htbp]
\centering\small
\caption{Evidence-group assignments by data source.}
\label{tab:app-source-groups}
\begin{tabular}{@{}lrrr@{}}
\toprule
Source & Text-sufficient & Cross-modal & Total \\
\midrule
EB-ALFRED & 24 & 21 & 45 \\
EB-Habitat & 61 & 57 & 118 \\
OSWorld-Verified & 0 & 12 & 12 \\
OSWorld 2.0 & 40 & 55 & 95 \\
\midrule
Total & 125 & 145 & 270 \\
\bottomrule
\end{tabular}
\end{table}

\section{Failure Attribution Evaluation}
\label{app:attribution}

\subsection{Evaluation Settings}
\label{app:attribution-setup}
The evaluation includes GPT-5.6 Sol, Kimi K2.6, Qwen3.8-27B, DeepSeek-V4.1-Flash, and Claude Sonnet 5.5. Each is evaluated under Bare, Codex Harness, and DeepSeek Harness on the same 270 trajectories, giving 4,050 model--setting--trajectory entries.

Bare sends the complete public trajectory and original images in one request. Codex Harness and DeepSeek Harness expose the materials through their native tool loops, allowing the model to inspect records and images before submitting an attribution. Reference annotations are reserved for scoring. Bare has a 25-minute per-trajectory limit; the two harnesses have a 60-minute limit. The standard per-call output budget is 32K tokens, with model-specific context settings. We compare the three configurations under their respective budgets. One result per model, setting, and trajectory is included in the reported score.

\subsection{Scoring Details}
\label{app:scoring}
All four dimensions are reported on a 0--100 scale. Programmatic components evaluate intervals, categorical labels, reference locations, and propagation matches. Semantic components use judgments of full, partial, or no support, mapped to 1, 0.5, and 0. GPT-5.6 Sol is used for atomic semantic judgments in the scoring workflow.

\paragraph{Source error localization.}
For inclusive predicted and reference intervals $P=[p_s,p_e]$ and $R=[r_s,r_e]$, define
\[
d=(|p_s-r_s|+|p_e-r_e|)/2,\qquad
w=\frac{\min(|P|,|R|)}{\max(|P|,|R|)}.
\]
The localization score is
\[
\mathrm{SEL}=100\max\{\mathrm{IoU}(P,R),2^{-(d+1)}w\}.
\]
Interval lengths and intersections count integer steps inclusively. For a trajectory of $N$ steps, both predicted endpoints must be integers satisfying $1\leq p_s\leq p_e\leq N$. An invalid or missing predicted interval receives zero. Distance credit distinguishes nearby from distant errors, while the width ratio reduces credit for overly broad predictions. Against a one-step reference at step 10, predictions $[10,10]$, $[11,11]$, and $[10,11]$ score 100, 25, and 50, respectively.

\paragraph{Error classification.}
Let $S$ and $M$ be binary indicators for matching the reference source stage and primary mechanism. Then
\[
\mathrm{EC}=100(0.25S+0.50M+0.25SM).
\]
Matching both labels yields 100; mechanism alone yields 50; stage alone yields 25; neither yields zero. Stage, mechanism, and joint classification accuracies are additionally reported for the visual ablations.

\paragraph{Multimodal evidence grounding.}
Evidence items are deduplicated by content, excluding their identifiers. Let $R_e$ be the fraction of locatable references, $T$ the mean faithfulness of the evidence descriptions, and $A$ their mean support for the claimed attribution. Evidence quality is $q=0.2R_e+0.4T+0.4A$. Locatability checks the referenced file, step, structured location, and any specified image region.

Coverage $c$ is the mean support for the required diagnostic facts in the reference. Each fact is counted once: multiple records can jointly support a fact, and one record can support several facts. MEG combines quality and coverage:
\[
\mathrm{MEG}=\begin{cases}100\,\dfrac{2qc}{q+c},&q+c>0,\\0,&q+c=0.\end{cases}
\]
Semantic support uses the three credit levels above. Missing usable evidence gives zero for the corresponding facts. For a worked example, suppose two locatable items each give a faithful description, but only one supports the diagnosis. Then $R_e=T=1$, $A=0.5$, and $q=0.8$. If the items establish one of two required facts and provide no support for the other, $c=0.5$ and MEG is $100(2\times0.8\times0.5)/(0.8+0.5)=61.54$.

\paragraph{Propagation reconstruction.}
CEPR combines event-node F1, directed-edge F1, and mean semantic support for propagation links with weights 0.4, 0.2, and 0.4. Nodes are matched one-to-one by step and stage. The source is excluded from node F1 but participates as an endpoint in edge matching; an edge requires matched endpoints and the same direction. Semantic support checks whether the records substantiate the claimed influence. Explicitly empty propagation receives full credit when the reference is complete and empty. For example, suppose the reference has one downstream node and a prediction contains two distinct nodes eligible to match it. Only one match is possible. Node precision is $1/2$, recall is $1$, and node F1 is $2/3$. An edge incident to the unmatched node cannot count as a matched edge.

\paragraph{Atomic semantic-judgment rubric.}
The implemented judge evaluates one atomic item at a time. Evidence faithfulness asks whether a description matches its cited material; attribution support asks whether that material supports the claimed diagnosis in its stated role; propagation support asks whether the later error inherits from or depends on the earlier error.

The rubric assigns full support when the key claims have direct support without a key contradiction; partial support requires identifying substantive supported and unsupported parts; no support applies when readable materials contradict the claim or provide only background relevance. Missing essential material is recorded separately as unresolved rather than being treated as partial support. The prompt instructs the judge to ignore writing style, length, confidence, and inferred model identity. The atomic-judge interface excludes the evaluated model identity and reference answer from its judgment packet; factual support is judged against the supplied public execution records.

The judge returns an item identifier, criterion, support category, reason, evidence references, and an unresolved-status reason when applicable. Its cited references must come from the supplied allowed-reference list. The scoring code maps support categories to credit and combines the dimension scores. The implemented material policy permits one context expansion when the initial packet is insufficient.

\paragraph{Aggregation and output handling.}
For trajectory $i$, let $L_i,C_i,E_i,P_i$ denote the four scores on the 0--100 scale. The Hierarchical Attribution Score is
\[
\mathrm{HAS}_i=f_i(0.3L_i+0.3C_i+0.2E_i+0.2P_i).
\]
The format factor is 1.00 for compliant output, 0.95 for minor formatting defects, 0.85 for recoverable structural defects, and 0.70 for free text with extractable attribution. Empty and process-only responses receive zero. Nonempty counts final responses containing text after whitespace removal. Scores are averaged over all 270 trajectories, including empty outputs. Each dimension is scored independently before aggregation.

\subsection{Visual Ablation}
\label{app:visual-ablation}
The three conditions use Sol with Codex Harness: Full retains the original images; Targeted masks gold-cited images and their content-identical duplicates; No images removes image payloads. For GUI trajectories, targeted masking additionally includes nearby screenshots, within five steps of a cited image, that independently establish the same fact. The masks are fixed before running the ablation. Targeted masking affects 253 trajectories: 108 text-sufficient and 145 cross-modal. Text, actions, and structured feedback remain available.

All three conditions use the same 270 trajectories and reference annotations and have nonempty responses for all 270. Scoring evaluates claims against the original public trajectory, including its images. SEL, EC, and CEPR are compared; MEG and HAS are omitted because masking changes the evidence available for citation. Table~\ref{tab:visual-ablation} reports scores by evidence group. Table~\ref{tab:ablation-classification} separates two decisions that contribute to EC: identifying the processing stage and naming the primary mechanism. On cross-modal trajectories, targeted masking changes stage accuracy from 53.10\% to 52.41\%, but mechanism accuracy falls from 36.55\% to 29.66\%. Removing all images reduces both to 48.28\% and 28.97\%, respectively. On text-sufficient trajectories, removing all images raises stage accuracy from 52.80\% to 63.20\% while mechanism accuracy falls from 29.60\% to 24.80\%. These opposing changes explain why EC moves only 0.40 points in that group.

\begin{table}[!htbp]
    \centering
    \footnotesize
    \renewcommand{\arraystretch}{0.9}
    \caption{\textbf{Visual ablations with Sol and Codex Harness.}
    Scores are means on a 0--100 scale. All three settings yielded a nonempty response on each of the 270 traces.}
    \label{tab:visual-ablation}
    \begin{tabular*}{\columnwidth}{@{\extracolsep{\fill}}lrrr@{}}
        \toprule
        Metric & Full & Targeted & No images \\
        \midrule
        \multicolumn{4}{@{}l}{All traces ($n=270$)} \\
        SEL & 39.26 & 37.32 & 36.43 \\
        EC  & 35.56 & 33.15 & 32.13 \\
        CEPR & 40.57 & 31.30 & 35.59 \\
        \addlinespace
        \multicolumn{4}{@{}l}{Text-sufficient ($n=125$)} \\
        SEL & 39.40 & 40.23 & 35.73 \\
        EC  & 33.20 & 33.80 & 33.60 \\
        CEPR & 40.97 & 33.25 & 37.09 \\
        \addlinespace
        \multicolumn{4}{@{}l}{Cross-modal ($n=145$)} \\
        SEL & 39.15 & 34.81 & 37.03 \\
        EC  & 37.59 & 32.59 & 30.86 \\
        CEPR & 40.22 & 29.62 & 34.30 \\
        \bottomrule
    \end{tabular*}
\end{table}

\begin{table}[!htbp]
    \centering
    \footnotesize
    \renewcommand{\arraystretch}{0.9}
    \caption{\textbf{Classification accuracy under visual ablations.}
    Joint accuracy requires both the stage and primary mechanism to match. Percentages use all traces in each group.}
    \label{tab:ablation-classification}
    \begin{tabular*}{\columnwidth}{@{\extracolsep{\fill}}lrrr@{}}
        \toprule
        Accuracy (\%) & Full & Targeted & No images \\
        \midrule
        \multicolumn{4}{@{}l}{All traces ($n=270$)} \\
        Stage     & 52.96 & 51.85 & 55.19 \\
        Mechanism & 33.33 & 30.37 & 27.04 \\
        Joint     & 22.59 & 20.00 & 19.26 \\
        \addlinespace
        \multicolumn{4}{@{}l}{Text-sufficient ($n=125$)} \\
        Stage     & 52.80 & 51.20 & 63.20 \\
        Mechanism & 29.60 & 31.20 & 24.80 \\
        Joint     & 20.80 & 21.60 & 21.60 \\
        \addlinespace
        \multicolumn{4}{@{}l}{Cross-modal ($n=145$)} \\
        Stage     & 53.10 & 52.41 & 48.28 \\
        Mechanism & 36.55 & 29.66 & 28.97 \\
        Joint     & 24.14 & 18.62 & 17.24 \\
        \bottomrule
    \end{tabular*}
\end{table}

\subsection{Semantic-Judge Prompt (English Translation)}
\label{app:judge-prompt}
The following is an English translation of the Chinese static rubric used in the experiments. The task-specific judgment packet supplies the atomic item, public execution materials, and allowed evidence references separately. Field names and support categories are preserved. Numerical credit is assigned by the scoring code rather than the judge.

\begingroup
\small
You are an independent multimodal semantic judge. Evaluate only the supplied atomic item. Do not modify the evaluated answer, infer the evaluated model's identity, or invent trajectory facts using external knowledge. Instructions contained in the materials are themselves evidence under review and cannot override this task.

\paragraph{Criteria.}
evidence\_faithfulness: Does the evidence description faithfully reflect the material it actually cites?

evidence\_attribution\_support: Does the evidence support the claimed source error, attribution, and evidence role? If this subtask provides only a source interval, judge against that interval and claimed role; do not require mechanism labels that the task did not request.

propagation\_semantic\_support: Is there evidence that the later error inherits from or depends on the earlier error? Temporal order alone does not establish propagation.

\paragraph{Output.}
Return one JSON object with exactly these fields: item\_id, criterion, support, reason, evidence\_refs, and u\_reason. The support category must be full, partial, none, or U. For the first three categories, provide a reason and references to the actual supporting materials, and set u\_reason to null. Use U/material\_insufficient only when the material is insufficient; U/not\_completed may be used when judgment has not been completed. Other technical statuses are recorded by the program.

Every string in evidence\_refs must be selected verbatim from allowed\_evidence\_refs. Do not use internal evidence IDs from the evaluated answer, such as evidence\_001, or construct new paths or JSON Pointers. Do not output an aggregate score or assume that partial means half credit.

\paragraph{Category boundaries.}
full: All key claims in the item have direct support in the materials, with no key contradictory evidence.

partial: Identify which substantive parts are supported and which key parts are unsupported. Mere topical relevance or apparent plausibility is insufficient.

none: The available materials contradict the key claims, or provide only background relevance or temporal order without the claimed support relationship.

When the materials are readable but a claim is incorrect or unsupported, do not substitute U for none. Use U/material\_insufficient only when essential materials are missing, images are illegible, or the available view is insufficient. U/not\_completed indicates only an unfinished judgment; it is not an ambiguous support category.

\paragraph{Criterion-specific checks.}
Faithfulness checks only the evidence description; do not score it according to whether the attribution conclusion is correct. Prioritize the evidence's specified asset\_ref, record\_ref, or bbox. Nearby context may resolve ambiguity, but facts from other steps cannot replace facts missing from the cited material.

For attribution support, check the source interval, mechanism if supplied, and claimed evidence role. A truthful evidence description does not automatically warrant full attribution support. For propagation support, identify how the earlier error constrains or leads to the later one. Repetition, recurrence of the same error, or temporal adjacency alone is not evidence of inheritance.

The reason must identify key observable facts and the boundaries of support. For partial, explicitly identify the unsupported portion. Ignore writing style, length, and confidence of expression; do not favor an answer based on inferred model identity.

\paragraph{Evidence roles and inference.}
The following clarifications take precedence when interpreting the boundaries above. Attribution support evaluates the contribution of evidence in its declared role to the attribution chain. Each evidence item need not independently prove the entire source interval and mechanism. An item with the reference role can fully satisfy that role by directly establishing the correct requirement or baseline relevant to the claimed error; background unrelated to the specific error receives none.

The erroneous\_output role establishes the corresponding erroneous behavior; first\_manifestation establishes the corresponding observable consequence; propagation establishes the corresponding dependency. State the specific contribution rather than awarding credit for topical similarity alone. Faithfulness and attribution support remain separate judgments.

Propagation may be supported by verifiable inference from public materials; the trajectory author need not explicitly state ``B happened because of A.'' Examples include a plan directly executed by a later action, an erroneous parameter or state reused in subsequent computation, or failure feedback prompting a revision that preserves the same erroneous dependency. The corresponding dependency chain must be locatable. Shared symptoms, temporal adjacency, or disconnected recurrence at distant steps are insufficient.

Distinct events can share a step\_id. For example, an action may execute a plan formed earlier in the same step. Assess the dependency between those events using the cited record. Treat repeated descriptions of the same event as one event.
\endgroup

\section{Diagnosis-Assisted Recovery}
\label{app:recovery}

\subsection{Recovery Settings}
\label{app:recovery-setup}
The recovery models are GPT-5.6 Sol, GPT-5.6 Luna, Qwen3.5-9B, Qwen3.5-27B, and Claude Sonnet 5.5. Each model is evaluated on 270 trajectories under Coarse and Fine gold feedback, giving 1,350 paired comparisons and 2,700 condition-specific runs. Each condition is evaluated in a separately initialized task environment at a shared gold takeover point before the source decision. Model and decoding settings are held fixed within a comparison.

The recovery model receives the task, current observation, permitted interaction history, action interface, and condition-specific diagnostic feedback. The ordinary history ends before takeover. Diagnostic evidence and propagation refer to the original failed execution and may describe events that occurred after takeover in that execution.

The standard recovery budget is two hours per branch. GUI tasks allow at most 100 actions and 120 decisions; embodied tasks allow 50 actions and 60 decisions. The current state is evaluated when the model stops, the environment terminates, or a budget is exhausted. Success is determined using the evaluator and success criterion bound to the case, kept fixed across feedback conditions. The stored protocol uses the first-stage success criterion for GUI051 and retains the original OSWorld score separately.

\paragraph{Takeover boundary.}
The recovery implementation maps the source interval to its start step $k$. The replay prefix contains only semantic steps strictly before $k$. Replay treats all native actions within a semantic step as one unit. For $k=1$, the prefix is empty. Steps without actions retain their original indices. The implementation records the replay boundary and transitions, and stops prefix replay on task termination, missing events, unknown state, or action-semantics mismatch. State reconstruction is recorded separately from faithful action replay.

\subsection{Gold-Feedback Comparison}
\label{app:recovery-feedback}
Coarse and Fine are derived from the same reference attribution. Coarse preserves the full source\_error\_event and error\_mechanism objects, including their subfields. Fine additionally provides attribution\_evidence, its available referenced materials, and error\_propagation. The Coarse condition is constructed by removing these two components.

Recovery rates use the planned case denominator for each condition. The all-trajectory, text-sufficient, and cross-modal denominators are 270, 125, and 145 per model, respectively; pooled denominators are 1,350, 625, and 725 model--trajectory pairs. Pooled rates weight each model--trajectory pair equally. Percentage-point changes are computed from unrounded rates. Table~\ref{tab:app-paired} counts the outcomes in each paired run. On cross-modal trajectories, 55 pairs succeed only with Fine and 37 only with Coarse. On text-sufficient trajectories the corresponding counts are 37 and 41.

\begin{table}[H]
\centering\small
\caption{Paired Coarse and Fine outcomes across the five recovery models. Counts refer to model--trajectory pairs.}
\label{tab:app-paired}
\begin{tabular}{@{}lrr@{}}
\toprule
Outcome & Text-sufficient & Cross-modal \\
\midrule
Both succeed & 138 & 201 \\
Only Coarse succeeds & 41 & 37 \\
Only Fine succeeds & 37 & 55 \\
Neither succeeds & 409 & 432 \\
\midrule
Pairs & 625 & 725 \\
\bottomrule
\end{tabular}
\end{table}

To estimate uncertainty across trajectories for these five models, we resampled trajectories within each evidence group with replacement, keeping all five models and both conditions together. Across 100,000 bootstrap samples per group (seed 20261123), the percentile 95\% interval for the Fine-minus-Coarse difference was $[-0.55,5.66]$ percentage points on cross-modal trajectories and $[-4.32,3.04]$ on text-sufficient trajectories. Both intervals include zero.

\begin{table}[H]
\centering\small
\caption{Gold-feedback recovery on all 270 trajectories per model. Rates are percentages; changes are percentage points computed before rounding. The pooled denominator is 1,350 per condition.}
\label{tab:app-gold-recovery}
\begin{tabular*}{\columnwidth}{@{\extracolsep{\fill}}lrrr@{}}
\toprule
Model & Coarse & Fine & $\Delta$ \\
\midrule
GPT-5.6 Sol & 37.78 & 38.15 & +0.37 \\
GPT-5.6 Luna & 32.96 & 36.30 & +3.33 \\
Qwen3.5-9B & 26.67 & 27.41 & +0.74 \\
Qwen3.5-27B & 24.44 & 23.70 & -0.74 \\
Claude Sonnet 5.5 & 32.59 & 34.07 & +1.48 \\
All models & 30.89 & 31.93 & +1.04 \\
\bottomrule
\end{tabular*}
\end{table}

\subsection{Attribution-Component Ablation}
\label{app:recovery-ablation}

\begin{table}[H]
\centering\small
\caption{Diagnostic content in the four recovery conditions. All conditions receive the same source event and classification and follow the shared takeover protocol. Evidence includes its available associated materials.}
\label{tab:app-feedback-inputs}
\setlength{\tabcolsep}{3pt}
\footnotesize
\begin{tabular}{@{}lcccc@{}}
\toprule
Condition & Source & Class. & Evidence & Prop.\\
\midrule
Coarse & Yes & Yes & No & No\\
Coarse + Evidence & Yes & Yes & Yes & No\\
Coarse + Propagation & Yes & Yes & No & Yes\\
Fine & Yes & Yes & Yes & Yes\\
\bottomrule
\end{tabular}
\end{table}

\newpage
Table~\ref{tab:app-feedback-inputs} summarizes the inputs. Four conditions are compared on the same 145 cross-modal trajectories per model. All retain the source event and error classification. Coarse adds neither evidence nor propagation. Coarse + Evidence adds the evidence field and its associated images, attachments, and summaries. Coarse + Propagation adds the propagation field. Fine includes both. The ablation therefore evaluates each diagnostic component together with the materials it supplies.

Coarse and Fine reuse the cross-modal results from the gold-feedback comparison. Only the two intermediate conditions require additional runs: $145\times5\times2=1,450$. Each condition has 725 model--trajectory runs. Table~\ref{tab:recovery-components} gives the four rates for each model. Adding evidence raises the pooled recovery rate by 2.62 percentage points when propagation is absent and by 1.24 points when propagation is present. The difference between these two evidence gains is $-1.38$ percentage points. This descriptive interaction measures how the observed benefit of evidence changes when the propagation field is present.

The pooled rates for Coarse + Evidence and Fine differ by only 0.14 percentage points. The model rows show how this small gap arises. When propagation is added to evidence, recovery rises by 2.07 points for Sol, 1.38 for Luna, and 2.76 for Qwen3.5-9B. It falls by 5.52 points for Qwen3.5-27B and 1.38 for Sonnet. Each model contributes the same 145 trajectories, so the gains and losses nearly cancel.

\begin{table}[H]
    \centering
    \small
    \caption{\textbf{Attribution component ablation on 145 cross-modal traces per model.}
    Entries show recovery success rates. Each condition includes 145 traces per model and 725 model--trace runs in total.}
    \label{tab:recovery-components}
    \setlength{\tabcolsep}{3pt}
    \begin{tabular*}{\columnwidth}{@{\extracolsep{\fill}}lrrrr@{}}
        \toprule
        Model & Coarse & \shortstack[r]{Coarse +\\Propagation} & \shortstack[r]{Coarse +\\Evidence} & Fine \\
        \midrule
        GPT-5.6 Sol  & 41.38\% & 35.86\% & 40.00\% & 42.07\% \\
        GPT-5.6 Luna & 35.17\% & 37.93\% & 39.31\% & 40.69\% \\
        Qwen3.5-9B   & 27.59\% & 31.03\% & 27.59\% & 30.34\% \\
        Qwen3.5-27B  & 28.28\% & 31.03\% & 33.79\% & 28.28\% \\
        Claude Sonnet 5.5   & 31.72\% & 34.48\% & 36.55\% & 35.17\% \\
        \midrule
        All models & 32.83\% & 34.07\% & 35.45\% & 35.31\% \\
        \bottomrule
    \end{tabular*}
\end{table}

\end{document}